\documentclass[preprint,5p,times,twocolumn]{elsarticle}

\usepackage{graphicx}
\usepackage{amssymb}

\usepackage{todonotes}

\usepackage{relsize}
\DeclareMathAlphabet\mathbfcal{OMS}{cmsy}{b}{n}
\usepackage{caption}
\usepackage{subcaption}
\usepackage{framed}
\usepackage{mathtools}
\DeclarePairedDelimiter\norm{\big\lVert}{\big\rVert}%
\DeclarePairedDelimiter\innerprod{\big\langle}{\big\rangle}%
\newcommand{\by}{\hspace{-0.5 pt}\mathsmaller{\times}\hspace{-0.5 pt}}
\newcommand{\Lsym}{\mathcal{L}^{\textit{\textsmaller{sym}}}}

\newcommand*{\pd}[3][]{\ensuremath{\frac{\partial^{#1} #2}{\partial #3}}}
\usepackage{algorithm,algpseudocode}
\usepackage{comment}

\usepackage{amsmath}

\newcommand{\bm}{\mathbf}

\makeatletter
\def\Let@{\def\\{\notag\math@cr}}
\makeatother

\journal{Pattern Recognition}

\begin{document}

\begin{frontmatter}


\title{Identifying noisy labels with a transductive semi-supervised leave-one-out filter}



\author{Bruno Klaus de Aquino Afonso} 

\cortext[cor1]{Corresponding author:}
\ead{bruno.klaus@unifesp.br}

\author{Lilian Berton} 
\address{Institute of Science and Technology -- Federal University of S\~ao Paulo
  (UNIFESP)\\
  S\~ao Jos\'e dos Campos -- SP -- Brazil}

\begin{abstract}
Obtaining data with meaningful labels is often costly and error-prone. In this situation, semi-supervised learning (SSL) approaches are interesting, as they leverage assumptions about the unlabeled data to make up for the limited amount of labels. However, in real-world situations, we cannot assume that the labeling process is infallible, and the accuracy of many SSL classifiers decreases significantly in the presence of label noise. In this work, we introduce the $\texttt{LGC\textunderscore LVOf}$, a leave-one-out filtering approach based on the Local and Global Consistency (LGC) algorithm. Our method aims to detect and remove wrong labels, and thus can be used as a preprocessing step to any SSL classifier. Given the propagation matrix, detecting noisy labels takes $\mathcal{O}(cl)$ per step, with $c$ the number of classes and $l$ the number of labels. Moreover, one does not need to compute the whole matrix, but only a $l \,\by\, l$ submatrix corresponding to interactions between labeled instances. As a result, our approach is best suited to datasets with a large amount of unlabeled data but not many labels. Results are provided for a number of datasets, including \texttt{MNIST} and \texttt{ISOLET}.  $\texttt{LGC\textunderscore LVOf}$ appears to be equally or more precise than the adapted gradient-based filter, and thus can be used in practice for active learning, where it may iteratively send labels for re-evaluation. We show that the best-case accuracy of the embedding of $\texttt{LGC\textunderscore LVOf}$ into LGC yields performance comparable to the best-case of $\ell_1$-based classifiers designed to be robust to label noise. 

\end{abstract}


\end{frontmatter}


\section{Introduction}
\label{S:1}
In recent years, many machine learning applications have relied on building increasingly complex models and tuning them with large datasets \cite{dobre2014}. A likely reason for this trend is the fact that it has become easier to get access to a significant volume of data quickly. The Internet of Things (IoT) has made it so that every kind of device is continuously connected, sending information through the network \cite{ahmed2017role}. Predictive models are able to recognize the patterns within this flux and utilize this knowledge to fulfill a multitude of tasks. 

Not all data are made equal. The central question about the data at hand is whether a label is present or not. By label, we mean some meaningful information that we want to predict new instances, as in a class that is assigned to some object. In the best-case scenario, all of our data is labeled and inference is based on generalizing from instances paired with their respective labels. As it turns out, obtaining labels is often a lengthy and costly process. For some specific datasets, such as the ones found in biological research, this activity requires individuals with enough expertise. 
Depending on the problem, it is simply infeasible to obtain all labels, and a compromise must be made.

Semi-Supervised Learning (SSL) addresses the problem of scarce labels \cite{Chapelle_etal_2010}. If the data distribution can tell us something about the label distribution, SSL should be able to leverage unlabeled data instead of discarding it. By doing so, it is possible to attain comparable classification performance at a much lower cost. For example, if classes are linked to well-separated clusters, SSL can identify them and use this information in future predictions. Despite relying on stronger assumptions about the data, SSL has been successfully employed for decades.
 Whenever labels are obtained as a result of a crowd-sourcing effort or scraped from social networks, we can expect a fair amount of labels to be incorrect. Research shows that, except for a handful of specific cases,  label noise reduces the accuracy of classifiers, and its impact is not negligible \cite{frenay2013classification}. 
 
 There are several ways to mitigate label noise. First of all, reducing model complexity can make the classifier robust by virtue of not picking a complex hypothesis just to fit a couple of incorrect labels. Another option is to modify some components of the algorithm. For example, the hinge loss of a support vector machine (SVM) is not robust even to uniform noise \cite{manwani2013noise}. Also, the learning procedure found in many boosting-based classifiers, such as Adaboost \cite{abellan2010bagging}, leaves them particularly vulnerable to label noise, as noisy instances are given progressively more importance later in training.
 
Within the vast literature of SSL, we have found that most SSL classifiers that were introduced to try to mitigate label noise explicitly are graph-based, although other approaches also do exist. That is not to say that all graph-based SSL approaches are well equipped to deal with label noise, however.  One of the most popular SSL algorithms, named \textit{Gaussian Fields and Harmonic Functions} (GFHF) \cite{alg_GFHF}, forces the classification of labeled instances to be the same as the possibly noisy input. The \textit{Local and Global Consistency} (LGC) \cite{alg_LGC} classifier, which is the starting point of our work, handles it better by introducing a hyperparameter that controls the trade-off between fitting the labels and being smooth concerning the graph. Using the random walk interpretation of the algorithm, one equates this to using a slower decay for the importance of vertices reached later in the walk, which mitigates the effect of a noisy label on its immediate neighbors. Despite this, there is much room for improvement. Namely, LGC does not explicitly try to address the noisy labels (i.e., by changing or removing them). One can configure the underlying diffusion process so that the noisy signals are mostly overwhelmed by correct signals, but ultimately it would be better if they were not there altogether.
Another problem is the use of the $\ell_2$ norm, which makes local differences on the graph that are either all small or all large and cannot be set to precisely zero \cite{alg_GTF}. As a result, several classifiers based on the $\ell_1$ norm have emerged \cite{alg_GTF,alg_LSSC,alg_SIIS}, and they have been shown to be more robust to label noise. It is also worth mentioning the concept of \textit{smooth eigenbasis pursuit} \cite{alg_SIIS}, which consists of restricting the output (or possibly the input) to a linear combination of vectors from a basis of smooth functions obtained from the decomposition of the graph Laplacian. This concept is found within many classifiers and provides a complementary way of regularizing the function. Finally, one of the approaches closest to what we have done is the \textit{Label Diagnosis through Self-Tuning (LDST)} \cite{alg_LDST}, which greedily modifies the labels by following the gradient of the LGC cost function.

\par Our approach, the \textbf{Leave-One-Out filter based on Local and Global Consistency \texttt{(LGC\textunderscore LVOf})}, takes an orthogonal approach to the ones described above. Whereas some filters may take the point of view of reconstructing from a sparse eigenbasis \cite{liu2011noise}, ours is based on detecting contradictions. We take advantage of the fact that the solution of the LGC algorithm takes the form $\mathbf{F = PY}$, with $\mathbf{Y}$ the binary matrix of initial labels and  $\mathbf{P}$ the \textbf{propagation matrix}. Notably,  $\mathbf{P}$ gives us the final labels as a linear function of the known initial labels, and $\mathbf{P}_{ij}$ is the influence instance $j$ has on $i$, as determined by the diffusion process. Computing the propagation matrix is often undesirable, for two reasons: first, the computation can be way slower than computing the result $\mathbf{F}$ directly; second, it is a dense  $n$ by $n$ matrix and memory runs out fairly quickly as the number $n$ of instances increases. We address both these issues. Assume a fixed number of classes and a constant amount of neighbors in the graph. \textbf{Our method needs only a $l$ by $l$ submatrix, which can be approximated iteratively, taking $\mathcal{O}(nl)$ time per iteration}. This means that, despite being ill-suited to the nearly supervised case ($l\!\approx\! n$), it works well when we have massive data with very few labels, which is when the semi-supervised paradigm is at its most useful. Finally, once we do have the propagation matrix available to us, detecting a noisy label is as fast as $\mathcal{O}(l)$ per label. Perhaps the most crucial detail is setting the \textbf{self-influence} (i.e., the diagonal) of the propagation matrix to zero. As a result, the i-th row of  $\mathbf{P}\mathbf{Y}$ contains the label assigned to instance $i$ if it was not considered in the diffusion process. This is then converted to class probabilities via row-normalization, and we compare the result of the modified diffusion to the initial label. The most contradicting one is assumed to be noisy, and we can remove (or correct) the effect that the label has on $\mathbf{F}$ efficiently.
\par For our experiments, we found out that, 
\begin{itemize}
\itemsep0em
\item The \texttt{LSDT} filter to remove labels must be adapted carefully, to avoid selecting instances of the same class.
\item  $\texttt{LGC\textunderscore LVOf}$'s performance is (at least) comparable to the adapted \texttt{LDST} filter, while taking up less memory. As such, it is more memory-efficient than LDST and may be executed on larger datasets.
\item  On a best-case analysis, embedding $\texttt{LGC\textunderscore LVOf}$ inside LGC is comparable to $\ell_1$-norm classifiers on \texttt{ISOLET}.
\item  Underestimating noisy labels still improves over the LGC baseline. We provide heuristics to choose a safe estimate.
\end{itemize}

Section \ref{sec:related_work} presents some related work. 
The necessary graph-based SSL concepts to comprehend the ideas behind $\texttt{LGC\textunderscore LVOf}$  are presented in section \ref{sec:theory}, with a special focus on LGC. Section \ref{sec:alg} presents the ideas and implementation of our filter, as well as a complexity analysis. Results comparing this filter to the adapted LDST gradient-based filter, the LGC without a filter, and robust $\ell_1$ methods are presented in section \ref{sec:experiments}. Finally, future work and concluding remarks are found in section \ref{sec:conclusion}.

\section{Related work} \label{sec:related_work}

The problem of learning with noisy labels in supervised learning has been well researched. According to the survey in  \cite{frenay2013classification}, there are a few different ways to deal with label noise.  It is known that the choice of learning model and loss function influences the effect of label noise, and those less affected by noise are said to be \textit{label noise-robust}. Those methods do not address the noise explicitly: they rely on overfitting avoidance to diminish its effect. On the other hand, \textit{label noise-tolerant} approaches are designed to consider label noise. That is, either modify the inner workings of a classifier to address label noise, or model the noise process directly with a probabilistic approach \cite{bhatia2015robust}. Finally, one can perform data cleansing, i.e. removing or relabeling a subset of labels identified as noisy. Our filter falls into this category. Data cleansing can also be a part of the learning algorithm.

Label noise has also been studied in SSL. The hinge loss of a Transductive Support Vector Machine is susceptible to noise, and better results have been achieved by changing it \cite{cevikalp2017large}. The co-training framework has also been adapted to consider label noise \cite{yu2011bayesian}. Graph-based SSL has had a considerable amount of different approaches to deal with few and noisy labels. Restricting the number of eigenfunctions can achieve this indirectly via overfitting avoidance \cite{Afonso_2020}. The self-paced learning paradigm has also been adapted to provide robustness \cite{gong2017learning}. Many different cost functions have been formulated, many of them making use of the $\ell_1$ norm \cite{alg_GTF,alg_LSSC,alg_SIIS}. Finally, there is the bivariate, gradient-based formulation \cite{alg_LDST}, with embedded filtering criteria that will be compared to ours in section \ref{sec:alg}. A comparison of a few traditional graph-based SSL classifiers subject to label noise is given by \citet{Afonso_2020}.

Noisy labeled samples can be a problem in many real-world applications. In \cite{Mozafari_2018}, the authors evaluated deep neural networks for classifying skin anomaly detection. They showed that some techniques for deep learning calibration do not work properly when the data set has a small size or contains samples with noisy labels. The Kaggle platform hosted a task ``Freesound Audio Tagging 2019'' with noisy labeled data and a smaller set of manually-labeled data \cite{Fonseca_2019}. In face recognition, there are some noisy face datasets and approaches to overcome this problem \cite{Wang_2019}. Usually, it is not the case that one has at hand both a set of clean and noisy labels; as such, for evaluation purposes, it is common in practice to take a dataset that was slowly and carefully labeled, assume that the available labels are clean, and corrupt a set amount of them.
\section{Theory}
In this section, we go over some necessary concepts  and set up the problem, moreover we review the well-known LGC classifier for semi-supervised classification.
\label{sec:theory}
\subsection{Assumptions and definitions}
\noindent The number of \textbf{instances} is $n$.
The number of \textbf{dimensions} of the input is $d$. The number of \textbf{classes} is $c$. The usual matrices for transductive graph-based algorithms are: the \textbf{input matrix}  $X \in M(\mathbb{R}_{n\by d})$; the \textbf{similarity matrix} $W \in M(\mathbb{R}_{n\by n})$; The  \textbf{label matrix} $Y \in M(\mathbb{R}_{n\by c})$;  the  \textbf{classification matrix} $F \in M(\mathbb{R}_{n\by c})$.
In this work, the weights of the affinity matrix are given by the Radial Basis Function (RBF) kernel. If $X_i,X_j$ are neighbors:
\begin{equation}
    W_{ij} =  exp\left(-\frac{\norm{X_i - X_j}^2}{2\sigma^2}\right)
\end{equation}
where $exp(x) = e^{x}$  is the exponential function. The similarity is at most 1, and decays like a gaussian with variance $\sigma^2$ as the distance increases.

The true label matrix $Y$ is such that $Y_{ij} = 1$ if the i-th instance has been labeled with the class indexed by $j$, and is zero everywhere else. Another very important matrix is the \textbf{degree matrix}  $D \in M_{nxn}$, a diagonal matrix whose entries are the row-wise sum of $W$.
It is also used to determine the \textbf{unnormalized graph Laplacian}
\begin{equation}
    L = D - W
\end{equation}
If W is symmetric, then it holds
\begin{align}
    (f^{\top}Lf) &= \frac12 \sum_i  \sum_j W_{ij} (f_i - f_j)^2
\end{align}
This last expression is also known as the \textbf{smoothness criterion} of this Laplacian, as it is a measure on how much the function changes on neighboring instances in the graph. 

The symmetrically normalized graph Laplacian is given by
\begin{equation}
     \Lsym = D^{-\frac12} L D^{-\frac12} = I - D^{-\frac12} W D^{-\frac12} 
\end{equation}
Note that:
\begin{align}
    f^{\top} \Lsym f = g^{\top} L g  = \frac12 \sum_j W_{ij} \left(\frac{f_i}{\sqrt{D_{ii}}} -  \frac{f_j}{\sqrt{D_{jj}}}  \right)^2\
\end{align}
where $g = D^{-\frac12}f$.

Unlike the un-normalized laplacian L, $\Lsym$ is necessarily \textbf{symmetric}. Thus, $\Lsym = Q \Lambda Q^{\top}$ for an orthogonal matrix $Q$. Each column of $Q$ corresponds to an eigenvector, and together they form an orthonormal basis for $\mathbb{R}^{n}$. $Q^{\top} f$ projects $f$ onto the eigenvector basis.

The symmetrically normalized laplacian \textbf{$\Lsym$ is a positive semi-definite matrix}. Let $g$ be an eigenvector of $\Lsym$, and $f = D^{\frac12}g$. Then, the corresponding eigenvalue $\lambda$ is given by:
\begin{equation}
    \lambda = \frac{\innerprod{g,\Lsym g}}{\innerprod{g,g}} = \frac{ \frac12 \sum_i  \sum_j W_{ij} (f_i - f_j)^2}{\sum_i D_{ii} f_i^2} \geq 0
\end{equation}
This tells us that eigenvectors with smaller eigenvalues are smoother with respect to the graph, and smooth functions will be projected by $Q$ mostly onto those eigenvectors.
\subsection{Using graph laplacians with multiple outputs}
The expression $f^{\top}Lf$ is a cost function related to the smoothness of $f$ with respect to the graph, given that the output of $f$ is a single scalar. Now consider the case where, instead of $f$, we have a matrix F such that $F_{ij}$ is proportional to the belief that instance $i$ should be assigned to class $j$. In this case, we must apply the graph laplacian to each column individually. The adapted smoothness cost is the sum:
\begin{equation}
    \sum_{k=1}^{c} (F_{[:,k]})L(F_{[:,k]})^{\top} = tr(F^{\top}LF)
\end{equation}
where $tr$ is the trace of the matrix.
\subsection{Revisiting Local and Global Consistency (LGC)}
The LGC algorithm  \cite{alg_LGC} is one of the most widely known graph-based semi-supervised algorithms. The cost being minimized is as follows: 
\begin{equation}
    \mathcal{Q}(F) = \frac12 \left(tr(F^{\top} \Lsym F) + \mu \norm{F - Y}^2 \right)
\end{equation}
The parameter $\mu \in (0, \infty)$ controls the trade-off between fitting labels versus enforcing the graph smoothness by minimizing local differences. 
By taking the derivative of $\mathcal{Q}$ with respect to $F$, one has
\begin{align}
\label{eq:der}
    \pd{\mathcal{Q}}{F} &=  \frac12 \pd{tr(F^{\top} \Lsym F)}{F} + \frac12 \mu\pd{ \norm{F - Y}^2}{F}\\
    &= ((1+\mu) I - S)F - \mu Y
\end{align}
with $S = D^{-\frac12}WD^{-\frac12} = I - \Lsym$. And, if we define $\alpha$ as
\begin{equation}
\alpha := 1/(1+\mu)   
\end{equation}
Then, by dividing Equation \ref{eq:der} by ($1+ \mu$), we observe that the derivative is zero exactly when
\begin{equation}
\label{eq:IminusalphaS}
    (I - \alpha S)F = (1-\alpha) Y
\end{equation}
The matrix $(I - \alpha S)$ can be shown to be positive-definite, and therefore invertible. We obtain the optimal classification as 
\begin{equation}
    F =  (1-\alpha)(I - \alpha S)^{-1}Y
\end{equation}
We hereafter refer to  $(I - \alpha S)^{-1}$ as the \textbf{propagation matrix} $P$. Each entry $P_{ij}$ represents the amount of label information from $X_j$ that $X_i$ inherits.  
It can be shown \cite{alg_LGC} that this matrix is a result of a diffusion process, which is calculated via iteration:
\begin{align}
    &F(0) = Y\\
    &F(t\!+\!1) = \alpha S F(t) + (1\!-\!\alpha)Y 
    \label{eqn:F_iter}
\end{align}
Moreover, the closed expression for $F$ at any iteration is
\begin{equation}
    F(t) = (\alpha S)^{t-1}Y + (1-\alpha)\sum_{i=0}^{t-1}(\alpha S)^{i}Y
\end{equation}
S is similar to $D^{-1}W$, whose eigenvalues are always in the range $[-1,1]$. This ensures the first term vanishes as $t$ grows larger, whereas the second term converges to $PY$. Consequently, $P$ can be characterized as
\begin{align}
    P &= (1-\alpha)\lim_{t \rightarrow \infty} \sum_{i=0}^{t}(\alpha S)^{i}\\
    &= (1-\alpha)\lim_{t \rightarrow \infty} \sum_{i=0}^{t} \alpha^i D^{\frac12}(D^{-1}W)^{i}D^{-\frac12}
    \label{eqn:P_iter}
\end{align}
The \textbf{probability transition matrix} $D^{-1}W$ makes it so we can interpret the process as a random walk. Let us imagine a particle walking through the graph according to the transition matrix. Assume it began at a labeled vertex $v_a$, and at step  $i$ it reaches a labeled vertex $v_b$, initially labeled with class $k$. When this happens, $v_a$ receives a confidence boost to class $k$. This boost is proportional to $\alpha^i$. This gives us a good intuition as to the role of $\alpha$. More precisely, the contribution of vertices found later in the random walk decays exponentially according to $\alpha^i$.

\subsection{Label Diagnosis through Self-Tuning (LDST)}
The LGC classifier induces a model such that every final label is a linear combination of the initial labels, with the weights determined by a random walk process. But the initial labels may not be correct, so it would be better if the label matrix $Y$ could change.  LDST \cite{alg_LDST} does this via bivariate formulation: given a currently optimal $F$, how can we change the label matrix such that the cost function decreases? This is done by calculating a matrix $Q$ which is the  gradient of $Q(F,Y)$ w.r.t. $Y$:
\begin{equation}
    Q = (\bm{P}^{\top} \, \bm{\mathcal{L}}^{sym} \,\bm{P} + \mu(\bm{P} - \bm{I})^2 )\bm{Y} = \bm{A}\bm{Y}
\end{equation}
We remove a label by finding the maximum $Q_{ij}$ with $(i,j)$  such that $X_i$ has the label of class $j$; similarly, we add labels by finding the minimum $Q_{ij}$ among all $i$ corresponding to unlabeled instances. The label matrix can also be normalized to give more confidence to labels in higher-degree vertices, as well as controlling the overall influence of each class. LDST computes the propagation matrix in its entirety. Even though $\bm{A}$ can be  reused, it needs the full propagation matrix. Calculating the inverse matrix takes $O(n^3)$ operations. It is also a dense matrix, and as a result there simply isn't enough memory when the dataset is large enough.

\subsection{Using different types of norms}

Most of the previous label propagation algorithms rely on variants of the same smoothness criterion $\bm{F}^{\top}\bm{L}\bm{F}$. It can be shown \cite{alg_GTF} that the generalized Laplacian smoother is in fact a kind of $\ell_2$-norm:
\begin{align}
    \bm{F}^{\top}\bm{L}^{(k+1)}\bm{F} = \norm{\Delta^{(k+1)}\bm{F}}^2_2
\end{align}
where $\Delta^{(k+1)}$ is a recursively defined graph difference operator. In particular,  $\Delta^{(1)}$ is the weighted,  oriented incidence matrix of the graph. As it is an $\ell_2$-norm, the graph Laplacian smoother cannot set any graph differences to be exactly zero. As a result, it does not have good local adaptativity. In contrast, the $\ell_1$-norm minimization allows for a classifier that is sensitive (wiggly) in some regions but also constant or smooth in others. \textit{Large-Scale sparse coding} (LSSC) \cite{alg_LSSC} uses the $\ell_1$-norm to transform noise-robust SSL into a generalized sparse coding problem. Specifically, they restrict $\bm{F} = \bm{U}_{\mathcal{L}_m} \bm{v}$ where $\bm{U}_{\mathcal{L}_m}$ is an $n\times m$ matrix whose columns are the $m$ eigenvectors with smallest eigenvalues. Thus, the solution must be a combination of the $m$ eigenfunctions of the graph Laplacian that are smoothest w.r.t the manifold. This results in a cost
\begin{align}
    Q(\bm{\bm{v}}) =  \norm{\Delta_\mathcal{L} \bm{U}_{\mathcal{L}_m}\bm{\bm{v}}}_1 + \mu \norm{\bm{U}_{\mathcal{L}_m} \bm{\bm{v}} - \bm{Y}}^2_2 
\end{align}
where $\Delta_\mathcal{L}$ is the normalized Laplacian version of the graph difference operator $\Delta$. Another approach, \textit{Semi-Supervised learning under Inadequate and Incorrect supervision} (SIIS)    \cite{gong2017learning} applies to the unnormalized Laplacian $\bm{L}$ ideas similar to the ones in LSSC. Let $\bm{U}$  be the matrix containing the eigenvectors. SIIS also restricts classifying function to be a combination of the $m$ smoothest eigenfunctions of the graph Laplacian. To do this, the matrix of eigenfunctions $\bm{U}$ is replaced with $\bm{U}_m$, which contains only the $m$ eigenfunctions with smallest eigenvectors. The corresponding diagonal matrix of eigenvalues is $\Sigma_m$. Both the first and second term are regularized with an $\ell_{2,1}$ norm.  Additionally, there is a third term to further encourage use of the smoothest eigenfunctions out of the available ones. We end up with:
\begin{align}\label{eqn:SIIS_cost}
     Q(\bm{v}) =  \norm{\Delta U_m\bm{v}}_{2,1}  + \mu_1 \norm{(\bm{U}\bm{v})_l - \bm{Y}_l}_{2,1} + \mu_2 tr(\bm{v}^{\top}\Sigma_m\bm{v})
\end{align}

\section{Leave-One-Out filter based on the LGC algorithm (LGC\textunderscore LVOf)}
\label{sec:alg}
The Local and Global Consistency (LGC) algorithm is still widely used. This is maybe due to being able to take advantage of the intrinsic geometric properties of the data manifold while providing a straightforward and elegant solution $F = PY$. The trade-off term $\mu$ (or $\alpha$, depending on the parameterization), should also make it somewhat resistant to label noise. By decreasing $\mu$, one should expect the solution to be less eager to fit noisy labels that would make the function less smooth concerning the underlying graph. However, it would be best if we had a mechanism that detects the individual noisy labels. These labels could then be removed, replaced, or sent to a specialist for a more thorough analysis.  In the following section, we develop an algorithm for the  \textbf{leave-one-out filter for the LGC algorithm (LGC\textunderscore LVOf)}, which aims to identify these labels correctly. The reason for this choice of filter is quite simple. For most classifiers, a leave-one-out filter would imply removing each labeled instance and training from scratch to get a prediction for that instance alone. However,  \textbf{for the LGC classifier, once we have the propagation matrix, the leave-one-out filter can be computed extremely fast}.
We start by assuming we have a set \texttt{labeledIndexes} containing the indices of the labeled instances. Recall that entry $\bm{P}_{ij}$ measures the class influence of $X_{j}$ on $X_{i}$. The first key insight is to set each diagonal entry $P_{ii}$ to zero. The i-th row of  $F = PY$ now contains the prediction for $X_i$ coming from a slightly different label propagation process, specifically one where we have unlabeled instance $X_i$ while keeping the remaining labels. The second insight is that, \textbf{because we only care about the rows of F corresponding to labeled instances, and because the rows of $\bm{Y}$ corresponding to unlabeled data are zero, we only need to compute an $\bm{l \by l}$ submatrix of $\bm{P}$, with $l$ the number of labels}. If we wish to use this submatrix, all the rows corresponding to unlabeled data are removed from $Y$ and $F$. This contrasts with the LDST gradient-filter as presented in \cite{alg_LDST}, which uses a full propagation matrix that will not scale to big datasets. The calculation of the submatrix is discussed in section \ref{subsec:complex}.  The next step is to follow some criteria to determine the label that is more likely to be noisy. Our approach is to interpret the output of label propagation as class probabilities. To do that, however, we must ensure that they sum up to 1, which is not usually the case. To fix this, let $\texttt{to\textunderscore class\textunderscore probabilities(F,Y)}: (M_{n \by c}(\mathbb{R}) \times M_{n \by c}(\mathbb{R})) \rightarrow M_{n \by c}(\mathbb{R})$ be such that
\begin{equation}
    \texttt{\small to\textunderscore class\textunderscore probabilities(F,Y)}_{ij} = \begin{cases}
    \frac{F_{ij}}{\sum\limits_{1\leq k \leq c}F_{ik}}  &\text{if $\sum\limits_{1\leq k \leq c}\!F_{ik} > 0 $}\\ 
    Y_{ij}  &\text{otherwise}
    \end{cases}
    \label{eq:class_prob}
\end{equation}
The \textbf{difference matrix} Q is simply
\begin{equation}
    Q = \texttt{to\textunderscore class\textunderscore probabilities(F,Y)} - Y
\end{equation}
We are only concerned with the rows of $Q$ that correspond to currently labeled instances. Assume that $X_i$ is any instance labeled with class $j$. Then, $Q_{ij} \leq 0$ and $-Q_{ij}$ measures how much one believes that $X_i$ has a noisy label. Also note that $Q_{ik} \geq 0$ for $k \neq j$. This means we can identify the label most likely to be noisy by finding indices $(i^{*},j^{*}) $ such that  $Q_{i^{*}j^{*}}$ is minimal among labeled instances, and suggest a new label by selecting class $k^*$ that maximizes $Q_{i^{*}k}$. The index $i^{*}$ is then removed from \texttt{labeledIndexes}. The previous contribution coming from $X_{i^{*}}$ must also be removed. This can be done by subtracting the $i^{*}$-th row of P from the $j^{*}$-th column of $F$. The algorithm will continue to remove instances until a stopping criterion is reached. We can set it to a fixed number of iterations $t$, as in LDST. However, we do not take for granted that a clean validation set is available; therefore we have developed heuristics from observed results (see section \ref{sec:comp_results}). Converting $F$ to class probabilities is crucial. If the division by the row sums were not to be performed, instances that are far away from any other could potentially be detected as noisy. This formulation also handles the particular case where an instance cannot reach any labels, returning a row of zeros, preventing them from being selected.

\begin{algorithm}[t]
    \caption{Leave-one-out filter for the LGC algorithm (LGC\textunderscore LVOf)}
    \label{lgc_lvo_alg}
    \hspace*{\algorithmicindent} \textbf{Input:}  \\Initial binary label matrix $Y \in M_{n \by c}(\{0,1\})$;
    \\Propagation submatrix $P \in M_{l \by l}(\mathbb{R})$ (see section \ref{subsec:complex}) ;\\
    Number of labeled instances $l$\\
 \hspace*{\algorithmicindent} \textbf{Output:} 
    \\ \texttt{noisyIndexes}: Indices of identified noisy labels.
    \\\texttt{suggestedClasses}: Suggested classes for noisy labels .
    \begin{algorithmic}[1] 
        \Procedure{LGC\textunderscore LVOf}{$Y,P,l$} 
        
        \State $\texttt{labeledIndexes} \gets \{1 \ldots l \}$
\For{$i \in \texttt{labeledIndexes} $}
            \State $P_{ii} \gets 0$ \Comment{remove label self-influence$\,\,$}
        \EndFor
        \State $Y \gets [Y_{ij}]_{i \in \,\texttt{labeledIndexes}} $

        \State $F\gets PY$
         \State $\texttt{noisyIndexes} \gets \{ \}$;$\,\texttt{suggestedClasses} \gets \{ \}$
            \While{stopping criteria not reached}  
                \State $Q \gets  \texttt{to\textunderscore class\textunderscore probabilities(F,Y)} - Y$
                \State $(i^*,j^*) \gets argmin_{(i: i \in \texttt{labeledIndexes}; j: 1 \leq j \leq c)} \, Q_{ij}$
                \State $k^* \gets argmax_{(k: 1 \leq k \leq c)} Q_{i^{*}k}$
                
                \State $\text{Remove }i^*\text{ from }\texttt{labeledIndexes}$
                \State $\text{Add }i^*\text{ to }\texttt{noisyIndexes}$
                \State $\text{Add }k*\text{ to }\texttt{suggestedClasses}$
                \State $F_{[:,j*]} \gets F_{[:,j^*]} - P_{[:,i^*]}$ \label{alg:line:contr} \Comment{remove contribution$\,\,$}
            \EndWhile\label{euclidendwhile}
            \State return \texttt{noisyIndexes},\texttt{suggestedClasses}
        \EndProcedure
    \end{algorithmic}
    \label{alg:LGCLVOF}
\end{algorithm}

\subsection{Time and space complexity}
\label{subsec:complex}
We divide the complexity analysis into two steps: first, obtaining the propagation matrix $\mathbf{P} = (I-\alpha S)^{-1}$, then detecting all noisy instances. The matrix inversion algorithm has a time complexity of $\mathcal{O}(n^3)$, where $n$ is the total number of instances. However, for this filter, \textbf{only a subset of the propagation matrix needs to be computed}. Specifically, if we calculate $\mathbf{P}_{ij}$ only for labeled instances $i$ and $j$, the result is the same as if we set the remaining entries to zero (this is due to the unlabeled rows of $Y$ being zero). Next, assume $\{X_1, \ldots, X_{l}\}$ are the labeled instances. The desired submatrix can be computed similarly to power iteration (Equation \ref{eqn:F_iter}), with the difference being that we substitute $Y$ by the following matrix:
\begin{equation}
    \widetilde{I} = \left[ \begin{array}{c}
I_{l\times l}  \\
\mathbf{0}_{(n-l)\times l} 
\end{array}
\right]
\end{equation}
This approach has two significant advantages. First, we avoid having to store a $n$ by $n$ matrix in memory, instead of settling for $l$ by $l$.  This is especially appealing within semi-supervised learning, as we usually expect $n \gg l$. Also, we can exploit sparsity by building our affinity matrix $W$ from a KNN graph (with time complexity of $\mathcal{O}(d n^2)$). By doing this, there will only be $kn$ nonzero entries for matrix $S$. Next, we compute a suitable approximation using $t$ iterations of (\ref{eqn:F_iter}), with complexity $\mathcal{O}(t\,knl)$. The analysis of the convergence properties related to label propagation is an extensive subject \cite{fujiwara2014efficient}, and we omit the discussion for brevity. With that in mind, we show that, when \textbf{P} is already computed, one can detect as many noisy instances as desired without exceeding $\mathcal{O}(cl^2)$ operations in total.  Before entering the \textbf{while} loop (line 10), the most expensive computation is the initial matrix multiplication that computes $F$, requiring $\mathcal{O}(c l^2)$ operations. The iterative process will require no more than $l$ steps, as that is the number of initially labeled instances. From within the loop, updating $Q$ and $F$ takes no more than $\mathcal{O}(cl)$ operations. In summary, this approach has a time complexity of: $\mathcal{O}(d n^2)$ for constructing the affinity matrix, $\,\mathcal{O}(ct\,knl)$ to approximate the propagation submatrix, and $\,\mathcal{O}(c l^2)$ to filter labels. Through the use of the submatrix, we store $\,\mathcal{O}((l+k)n)$ nonzero entries.

\section{Experiments}
\label{sec:experiments}
In this section, we go over our experimental setup, the chosen datasets and baselines, and discuss the obtained results.

\subsection{Datasets}
The datasets we chose can be roughly divided into two sets. To compare with LDST, we chose some known small-sized benchmarks, such that the full propagation matrix could be computed without compromises. On the other hand, we also selected a few larger datasets to analyze our filter with $\ell_1$-norm methods.
\paragraph{Chapelle's benchmarks: Digit1, $COIL_2$ and USPS}
These three datasets are benchmarks associated with the book \cite{Chapelle_etal_2010}  \footnote{Chapelle's Digit1, USPS, COIL available on http://olivier.chapelle.cc/ssl-book/benchmarks.html}. The latter two are modified versions of datasets found elsewhere, made more challenging by partially obscuring data.  The \textit{Digit1} dataset consists of a series of artificially generated images of the digit 1.  Downsampling and omission of certain pixels reduce it to a 241-dimensional instance. The two classes are created from thresholding the tilt angle. 
The \textit{Columbia object image library (COIL-100)} \cite{nene1996columbia} is a set of pictures taken of a hundred different objects from different angles, using steps of 5 degrees. $COIL_2$ is a slightly different version, as it only has two classes and is restricted to the red channel of 24 objects. The \textit{USPS} dataset is based on (but not identical to) the popular USPS handwritten digits. In this version, each digit is represented by 150 images. Digits 2 and 5 are given the same class, the remaining digits are assigned the other. As a result, this dataset is imbalanced.

\paragraph{MNIST}
This is a handwritten digit dataset based on the Modified National Institute of Standards and Technology database \cite{lecun1998gradient}, comprising 70000 grayscale images, each with a height and width of 28 pixels, for a total input dimension of 784. Each digit from 0 to 9 is represented. The frequency of digits is nearly balanced, as each class has a number of representatives in the range of  6 to 8 thousand.

\paragraph{ISOLET}
The Isolated Letter Speech Recognition \cite{fanty1991spoken} is a dataset related to audio data. To create this dataset, 150 subjects had to pronounce each of the 26 letters of the alphabet twice. The actual number of instances is $7797$, due to 3 missing examples. There are 26 classes, which identify the correct letter being uttered within the range A-Z.
\subsection{Experimental setup}
The programming language of choice was Python\footnote{code soon available on the author's GitHub: https://github.com/brunoklaus}. We manually implemented LGC, $\texttt{LGC\textunderscore LVOf}$, LDST and SIIS. We used a GPU to  speed up computation. 
\paragraph{Shared settings}
The affinity matrix is constructed with a symmetric KNN graph (there is an edge if either instance has the other within its neighbors) with $k = 15$ and an RBF kernel with $\sigma$ set heuristically to the mean distance to the 10th neighbor, as in \cite{Chapelle_etal_2010}; $\alpha \in \{0.99,0.9\}$, which are values used previously \cite{alg_LGC}\cite{Afonso_2020} that produce a good trade-off between prioritizing closer instances and taking more labels into consideration. Both LDST and our filter are given a hyperparameter $t$ to control the number of removed labels.  We make a point to \textbf{analyse the change in performance as we increase the number of instances removed}. There are 20 seeds used to randomly sample the labels and corrupt (i.e. flip to a different class) a percentage of them. 
\paragraph{Evaluation}
When comparing two filters, maintaining a high precision is the most important thing, especially when labels are few. As such, we plot the accuracy w.r.t. number of removed instances. When comparing our approach to a classifier, we must embed \texttt{LGC\textunderscore LVOf} into LGC and compare for accuracy. Validation is exceptionally difficult if we have few and noisy labels, so we compare the best-case scenario, as in \cite{alg_SIIS}. We compensate by analysing parametric sensitivity separately.
\paragraph{Experiment 1 - smaller datasets}
On smaller datasets, we have enough memory to compute the whole matrix and compare LDST with our filter. It is desirable for a filter to remove the most noisy instances while also maintaining high precision. As such, we illustrate the average precision of the filter as labels get removed. For all datasets, we repeatedly picked 150 labels and corrupted 30 (i.e. 20\%) of those.
\paragraph{Experiment 2 - larger datasets}
On larger datasets, we can't use LDST, so we compare our filter with $\ell_1$-norm methods instead. We compute the accuracy separately for labeled instances and unlabeled instances and provide the standard deviation. The experiment with the \texttt{ISOLET} dataset has a relatively large number of available labels (1040 out of 7797). This was done so that we could compare with the best-case results reported by \cite{alg_SIIS}.  We use 100 labels out of 70 thousand in \texttt{MNIST}. To compensate for this best-case scenario, the second part of this experiment shows the effect of overestimating or underestimating the amount of noisy labels. In this second part, we corrupt 30\% of labels in \texttt{MNIST}, and 20\% of \texttt{ISOLET}.

\begin{figure}[!t]
    \centering
\begin{subfigure}{0.95\linewidth}
    \includegraphics[trim={0 0 0 8.5cm},clip,width=\textwidth]{\detokenize{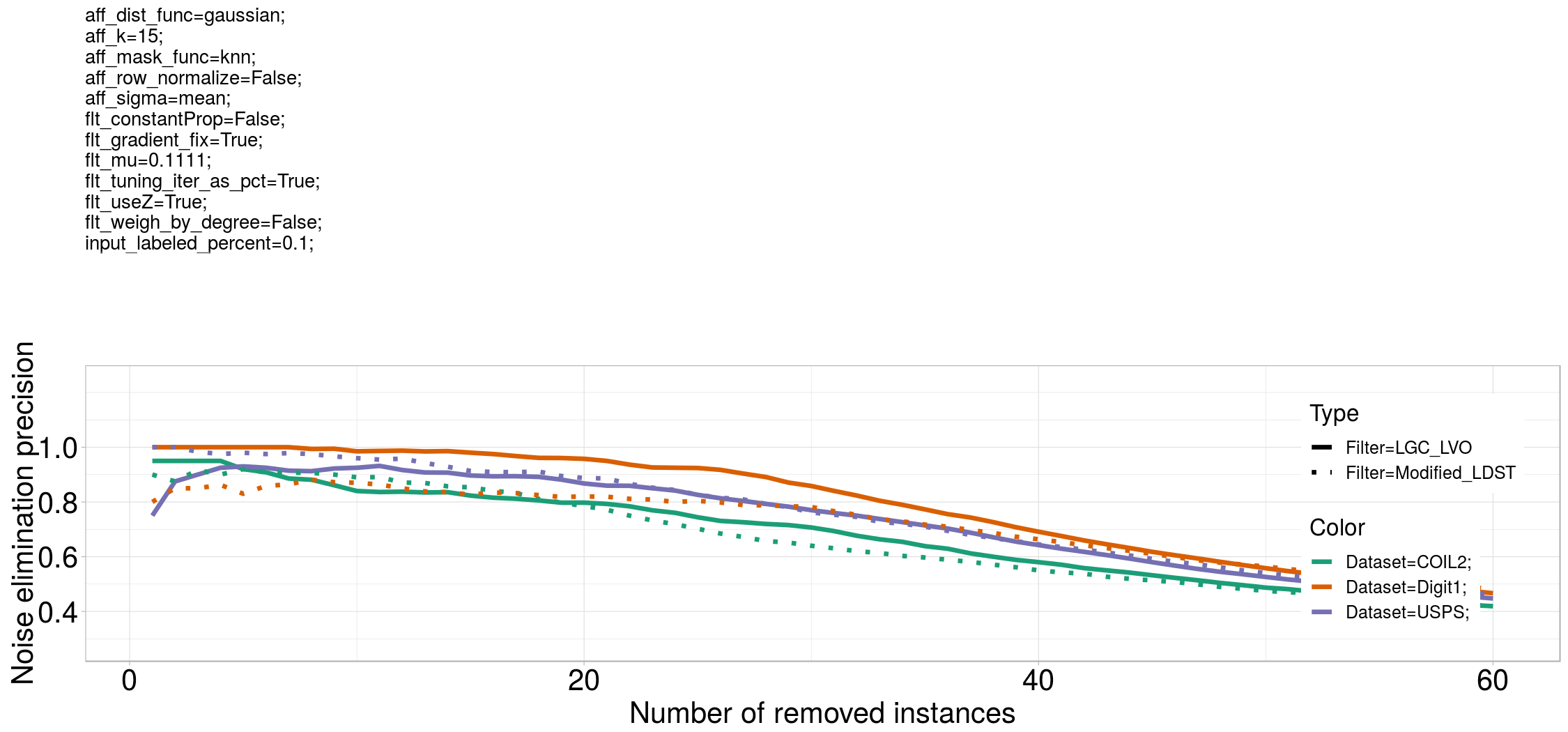}}
    \caption{$\alpha = 0.9$ (i.e., $\mu = 0.1111$)}
    
    \label{fig:prec_chap_1}
    
\end{subfigure}%
\\
\begin{subfigure}{0.95\linewidth}
    \includegraphics[trim={0 0 0 7.5cm},clip,width=\textwidth]{\detokenize{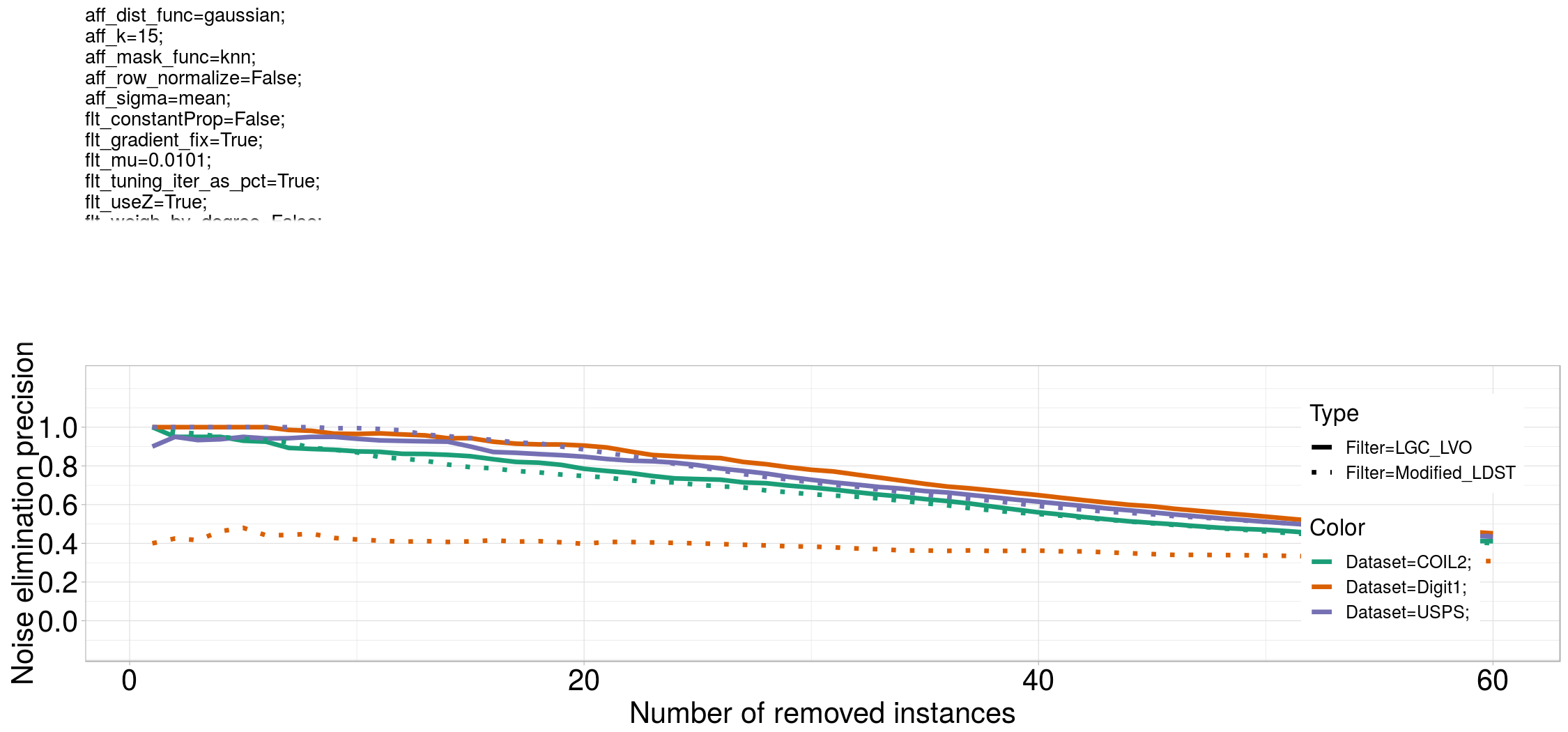}}
    \caption{$\alpha = 0.99$ (i.e., $\mu = 0.1010$)}
    
    \label{fig:prec_chap_2}
\end{subfigure}
    \caption{\texttt{COIL2,Digit1,USPS (30 noisy labels out of 150)}: Precision of LGCLVOf and Modified LDST }
\end{figure}

\begin{table}[!h]
\begin{subtable}{0.5\textwidth}
\centering
\scriptsize
\caption{Accuracy on unlabeled examples only}

\begin{tabular}{lllll}
\hline
Dataset & Noise Level & LSSC           & GTF                    & SIIS          \\\hline
                            & 0\%         & 84.8 $\pm$ 0.0   & 70.1 $\pm$ 0.0           & \bf{85.4 $\pm$ 0.0}  \\
\hfil ISOLET                & 20\%        & 82.8 $\pm$ 0.3   & 69.9 $\pm$ 0.2           & \bf{84.9 $\pm$ 0.6}  \\
\hfil (1040/7797 labels)    & 40\%        & 78.5 $\pm$ 0.6   & 59.8 $\pm$ 0.3           & \bf{80.2 $\pm$ 0.3}  \\
\hfil reported results      & 60 \%       & 67.5 $\pm$ 1.8   & 54.8 $\pm$ 0.5           & \bf{74.9 $\pm$ 0.14} \\ \hline
Dataset & Noise Level & LGC            & LGC\textunderscore LVOf &  SIIS\\    \hline
                            & 0\%         & 85.00 $\pm$ 0.61  & 85.00 $\pm$ 0.61     &  \bf{85.24 $\pm$ 0.32}             \\
\hfil ISOLET                & 20\%        & 82.89 $\pm$ 0.59  & 83.07 $\pm$ 0.65     &  \bf{83.69 $\pm$ 0.33}            \\
\hfil (1040/7797 labels)    & 40\%        & 79.33 $\pm$ 0.92  & 80.06 $\pm$ 1.01     &  \bf{80.88 $\pm$ 0.77}          \\
\hfil our results     & 60\%        & 70.69 $\pm$ 1.01  & \bf{74.37 $\pm$ 1.20}     &  72.97 $\pm$ 1.16
\end{tabular}
\label{tbl:isolet_unlabeled_acc}
\end{subtable}
\begin{subtable}{0.5\textwidth}
\tiny
\caption{Accuracy on labeled examples after label correction}
\centering

\scriptsize
\begin{tabular}{lllll}
\hline
Dataset & Noise Level & LSSC           & GTF                    & SIIS          \\\hline
                                & 0\%         & 89.9 $\pm$ 0.0   & \bf{95.8 $\pm$ 0.0}           & 91.1 $\pm$ 0.0  \\
\hfil ISOLET                    & 20\%        & 87.7 $\pm$ 0.3   & 79.8 $\pm$ 0.7           & \bf{90.5 $\pm$ 0.8}  \\
\hfil (1040/7797 labels)        & 40\%        & 82.9 $\pm$ 0.9   & 63.3 $\pm$ 0.4           & \bf{83.6 $\pm$ 1.0}  \\
\hfil reported results                & 60 \%       & 71.8 $\pm$ 1.7   & 55.3 $\pm$ 0.6           & \bf{77.4 $\pm$ 1.0} \\ \hline
Dataset & Noise Level & LGC            & LGC\textunderscore LVOf & SIIS       \\    \hline
                                & 0\%         & \bf{99.9 $\pm$ 0.02}   &     \bf{99.9 $\pm$ 0.02}   & 90.24 $\pm$ 0.69      \\
\hfil ISOLET                    & 20\%        & 80.84 $\pm$ 0.27     &       \bf{93.96 $\pm$ 0.88}  & 88.5 $\pm$ 1.07  \\
\hfil (1040/7797 labels)        & 40\%        & 60.24 $\pm$ 0.20    &      \bf{87.59 $\pm$ 1.02}    & 85.25 $\pm$ 0.96    \\
\hfil our results                     & 60\%        & 40.00 $\pm$ 0.04    &     \bf{78.52 $\pm$ 1.07}     & 76.34 $\pm$ 1.53   
\end{tabular}
\label{tbl:isolet_labeled_acc}
\end{subtable}
\caption{Best-case accuracy on \texttt{ISOLET} dataset}
\end{table}
\subsection{Adapting the LDST filter}
\label{subsec:adapt}
The LDST is defined in such a way that, at each step, one label is added and another is removed. We wanted to adapt this so that labels are only removed, much like our filter. If we simply remove the labeling operations, in practice this favored the removal of instances of the same class repeatedly. We managed to get better results by instead searching $(i,j)$ that \textit{maximizes} the gradient matrix $Q$, on the condition that $Y_{i\hat{\jmath}} = 1$ for $\hat{\jmath} \neq j$.

\subsection{Baselines}
We considered:  the modified LDST filter (as in section \ref{subsec:adapt}); the LGC classifier \cite{alg_LGC}; Large-Scale Sparse Coding (LSSC) \cite{alg_LSSC}; Graph Trend Filtering (GTF) \cite{alg_GTF}; Semi-Supervised Learning under Inadequate and Incorrect Supervision (SIIS) \cite{alg_SIIS}.

\begin{figure}[!t]
    \centering
\begin{subfigure}{\linewidth}
    \includegraphics[trim={0 0 0 8.3cm},clip,width=\textwidth]{\detokenize{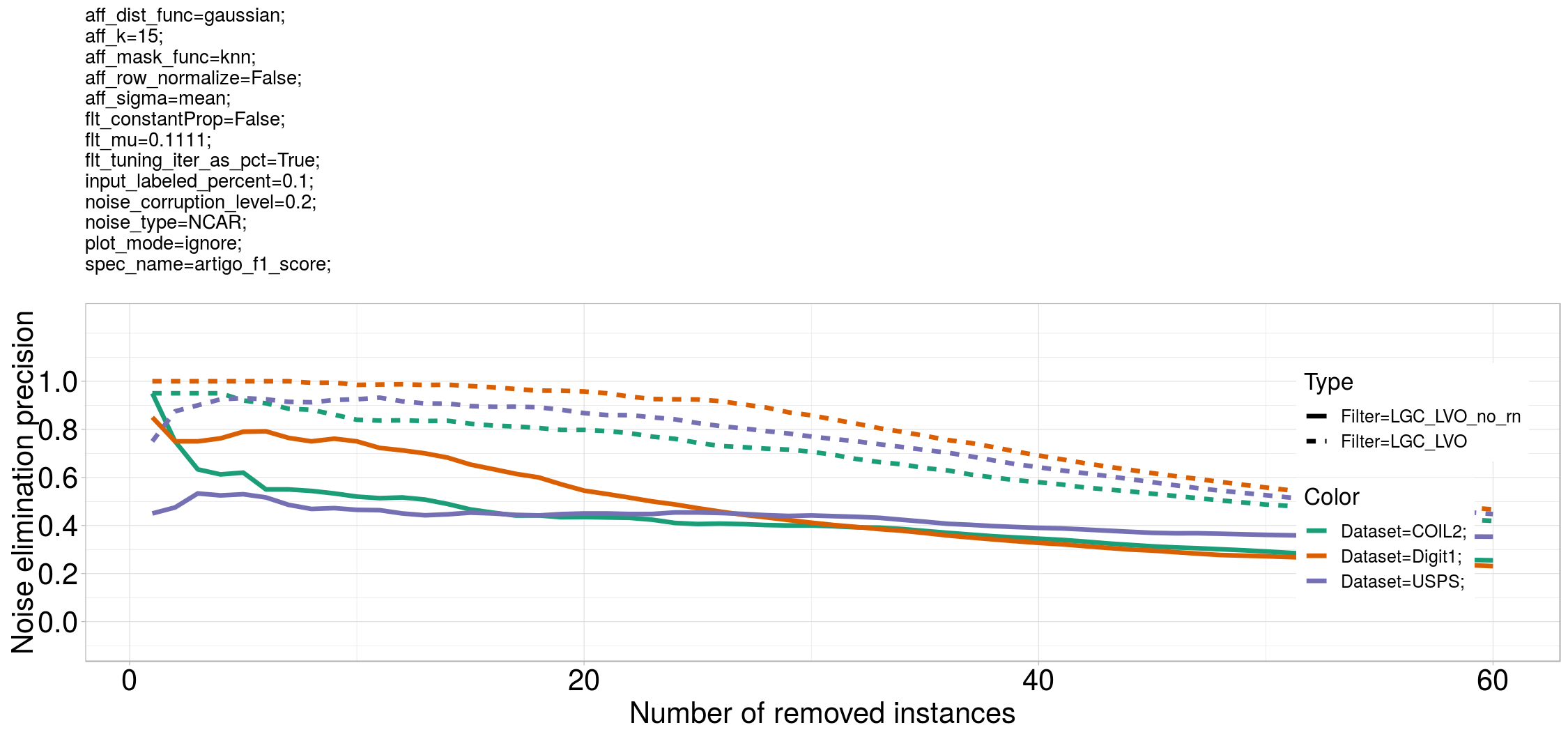}}
\end{subfigure}%

    \caption{Precision of \texttt{LGC\textunderscore LVOf} with and without row normalization (dashed and solid lines, respectively)}
    
    \label{fig:rn_chap_1}
\end{figure}
\subsection{Comparison results}
\label{sec:comp_results}

\paragraph{Experiment 1, $\texttt{COIL}_2$, \texttt{Digit1}, \texttt{USPS} }
We investigated the quality of our filter as the number of iterations (and consequently the number of removed instances) increases. This can be observed in Figures \ref{fig:prec_chap_1} and \ref{fig:prec_chap_2}. With $\alpha=0.9$, our filter was able to maintain 80\% precision on average while recalling 88.5\%, 72.3\%, 48.5\% of noisy labels respectively for \texttt{Digit1},\texttt{USPS}, and $\texttt{COIL}_2$.This order is expected, as Digit1 is a simple artificial dataset, whereas the increased amount of clusters of $\texttt{COIL}_2$ makes the task harder as the number of labels decreases. For the \texttt{Modified LDST}, these values were 64\%, 72.8\%, 48.6\%. In general, the \texttt{Modified LDST} and \texttt{LGC\textunderscore LVO} filters behaved similarly. When $\alpha = 0.99$, the modified LDST filter performed very poorly on Digit1 (after some more experiments, this only appeared to happen when $\mu$ was small). Aside from that, the observed precision changed little for both filters. Finally, on all datasets, we also observed a significant effect when row normalization was not performed (Figure \ref{fig:rn_chap_1}).
\paragraph{Experiment 2, MNIST and ISOLET}

\begin{figure}[!htb]
    \centering
    \begin{subfigure}{0.4\textwidth}
    \includegraphics[trim={0 0 0 11.5cm},clip,width=\textwidth]{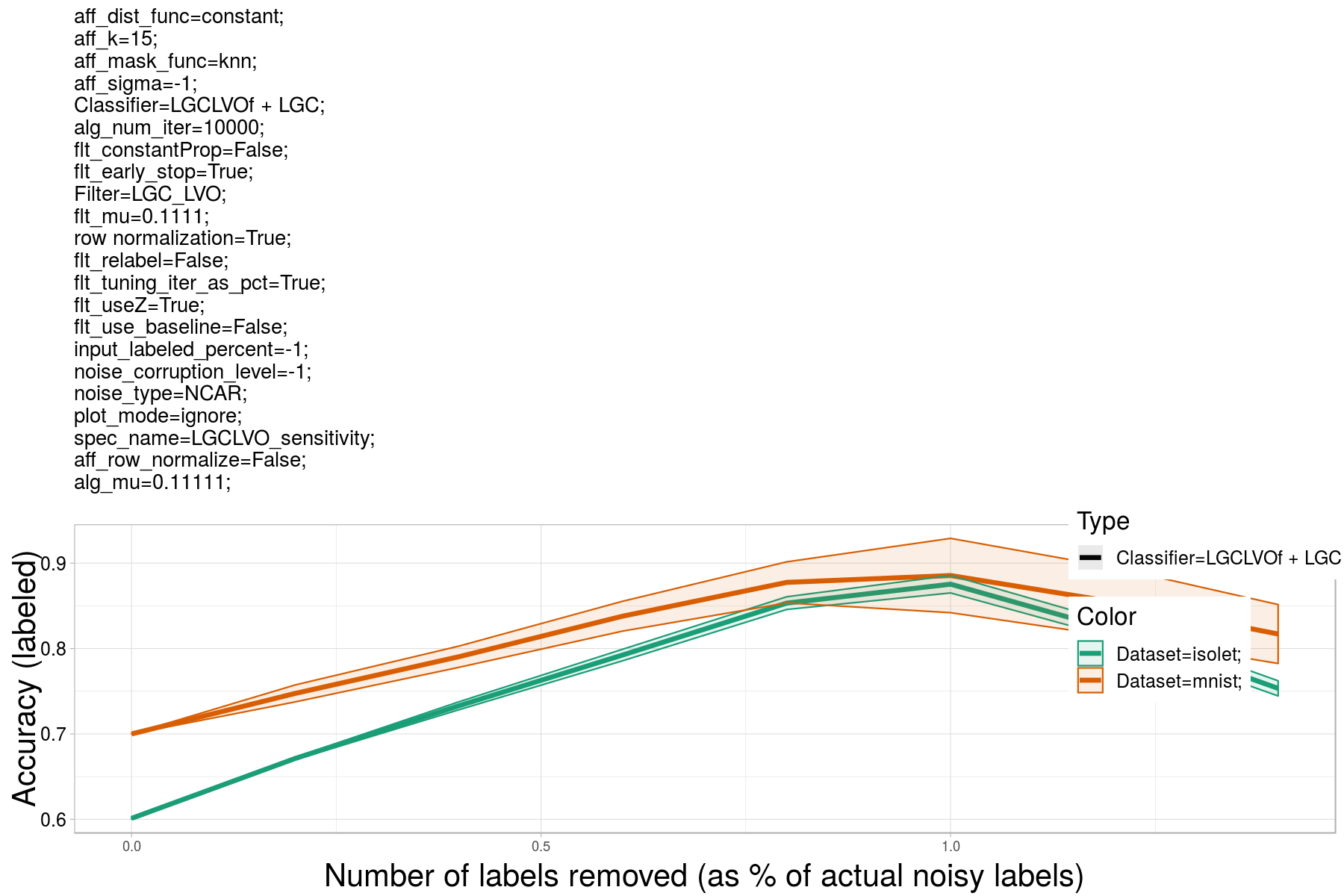}
    \end{subfigure}
    \begin{subfigure}{0.4\textwidth}
    \includegraphics[trim={0 0 0 11.5cm},clip,width=\textwidth]{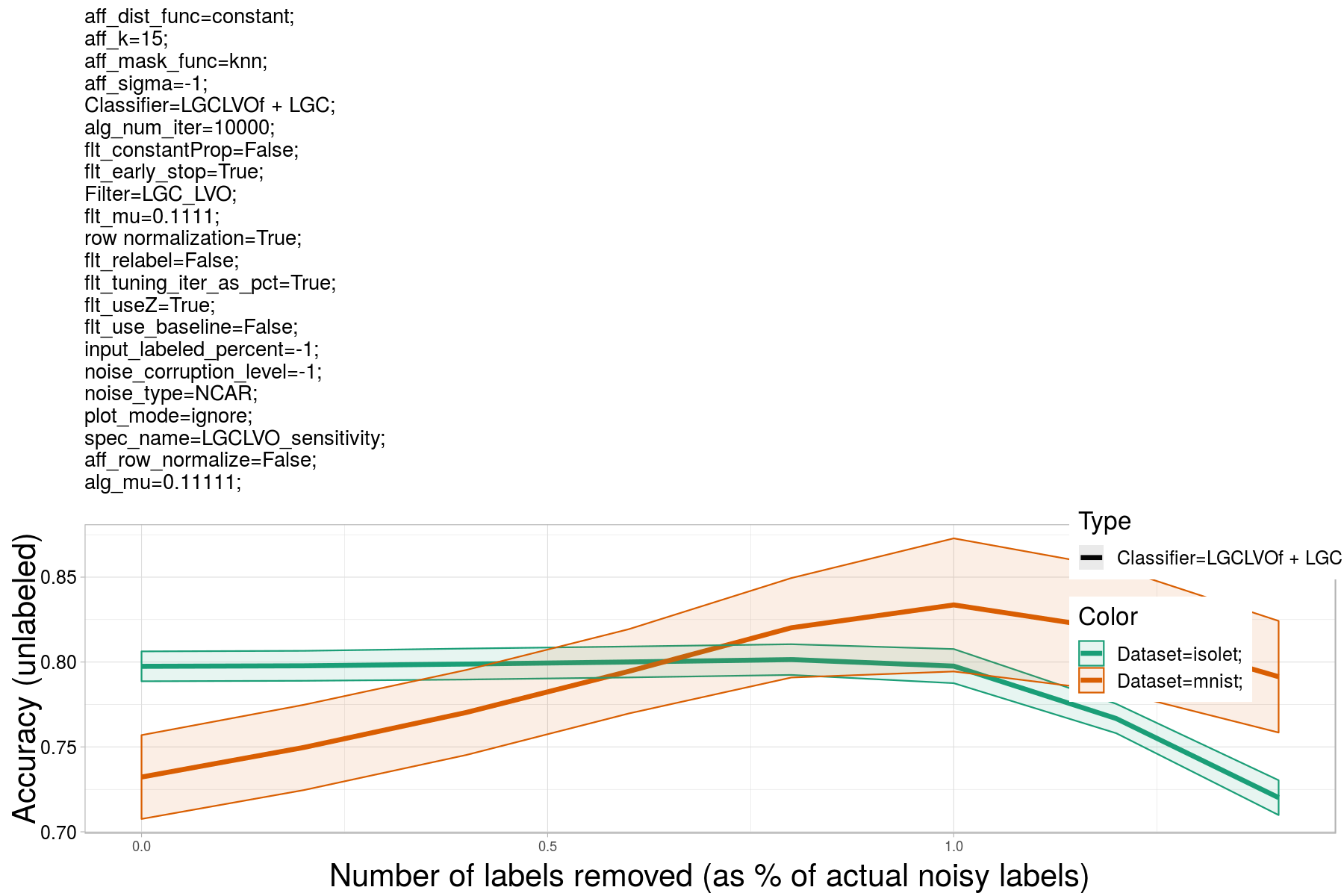}
    \end{subfigure}
    \caption{Accuracy w.r.t. the number of removed labels, for both labeled instances after correction, and unlabeled ones. Shaded regions indicate standard deviation.}
    \label{fig:LGCLVOf_sens}
\end{figure}
The MNIST dataset was too large for computing the entire propagation matrix, so we did not compare our filter with the \texttt{Modified LDST}. We were also unable to run our implementation of SIIS on MNIST. For \texttt{ISOLET}, Tables \ref{tbl:isolet_labeled_acc} and  \ref{tbl:isolet_unlabeled_acc} show the comparison of the best-case of our model against the previously reported results\cite{alg_SIIS}. For a fair comparison, our implementation of SIIS uses the affinity matrix as described in \cite{alg_SIIS}, and the same parameters. The best parameter for each noise level appeared to be consistent with previous results, as well as the decrease on accuracy, albeit slightly different because \cite{alg_SIIS} kept the label set fixed prior to corruption. For \texttt{LGC\textunderscore LVO}, we simply set $\alpha=0.9$ and performed a number of iterations equal to the amount of noisy labels, which should be a best-case scenario if the filter is able to maintain its precision. Having a perfect estimate on the amount of noisy labels is unrealistic. As Figure \ref{fig:LGCLVOf_sens} shows, underestimating 
this number has an effect on accuracy but generally  still yields some improvement over the LGC baseline. Overestimating it could be more dangerous, so in practice we recommend to use a conservative upper bound. One can also see that the LGC manages to be robust to label noise for \texttt{ISOLET} specifically. We believe that this is due to having more labels than usual, and that uniform label noise is more easily identifiable when you have many and well-separated classes. In spite of that, \texttt{LGC\textunderscore LVO} was better at correcting the noisy labels.
\begin{figure}[!t]
    \centering
\begin{subfigure}{0.5\linewidth}
    \includegraphics[trim={0 0 0 0.5cm},clip,width=\textwidth]{\detokenize{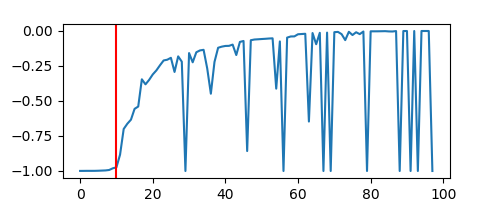}}
    
    \label{fig:f1}
    
\end{subfigure}%
\begin{subfigure}{0.5\linewidth}
    \includegraphics[trim={0 0 0 0.5cm},clip,width=\textwidth]{\detokenize{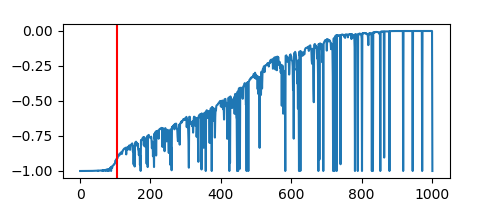}}

    \label{fig:f2}
\end{subfigure}
    \caption{Q values of MNIST and ISOLET. The red line is the number of noisy labels}
\label{fig:q_values}
\end{figure}
\paragraph{Guidelines}
\texttt{LGC\textunderscore LVOf} works best if there's much more unlabeled data than labeled data. A clean label set may or may not be available for validation. If it isn't, we recommend looking at the values from matrix $Q$. Those values measure how much the modified label propagation changed the original prediction. In MNIST and ISOLET, noisy labels corresponded to an initial plateau in Q values (Figure \ref{fig:q_values}). When in doubt, we recommend a conservative threshold (such as  $-0.8$), as it likely won't overestimate if the clean data complies with GSSL assumptions. 
\section{Concluding remarks and future work}
We have shown that the $\texttt{LGC\textunderscore LVO}$ filter can be an effective tool for detecting noisy labels within a semi-supervised task. Whereas many approaches throw out the $\ell_2$-norm due to its lack of robustness to noisy labels, our approach keeps it and explicitly addresses the noisy labels by detecting contradictions in the propagation model. This filter is comparable to LDST but more memory-efficient, as it works with a subset of the propagation matrix corresponding to interactions between labeled instances.
\par For future work, we aim to find a more elaborate criterion that gives us a reasonable estimate of the number of noisy labels. This may be done via an alternative Bayesian formulation that models the noise process directly. Moreover, if labels are exceptionally few, our previous work \cite{Afonso_2020} motivates a filter based on restricting eigenfunctions outright.

\label{sec:conclusion}

\section{Acknowledgements}
\label{sec:acknowledgements}
\footnotesize
This study was financed in part by the Coordination of Superior Level Staff Improvement (CAPES) - Finance Code 001 and Sao Paulo Research Foundation (FAPESP) grants \#2018/15014-0 and \#2018/01722-3.





\bibliographystyle{model1-num-names}
\bibliography{sample.bib}







\end{document}